\documentclass[letterpaper, 10 pt, conference]{ieeeconf}
\IEEEoverridecommandlockouts
\usepackage{caption}
\usepackage{graphicx}
\usepackage{float}
\usepackage{subcaption}
\usepackage{amsmath}
\usepackage{amssymb}
\usepackage[hidelinks]{hyperref}
\hypersetup{
  pdftitle={Intent-Handover: Grounding Language in Human-Usage Regions for Trustworthy Robot-to-Human Handovers},
  pdfauthor={Hanxin Zhang, Abdulqader Dhafer, Hongbiao Dong, Zhou Daniel Hao}
}
\usepackage{pgfplots}
\usepackage{makecell}
\usepackage{booktabs}
\usepackage[ruled,vlined]{algorithm2e}
\usetikzlibrary{patterns,calc}

\PassOptionsToPackage{shortlabels}{enumitem}
\let\labelindent\relax                      
\usepackage{enumitem}   

\pgfplotsset{compat=1.18} 

\title{\LARGE \bf
Intent-Handover: Grounding Language in Human-Usage Regions for Trustworthy Robot-to-Human Handovers
}

\author{Hanxin Zhang$^{1,2}$, Abdulqader Dhafer$^{1,2}$, Hongbiao Dong$^{3}$, Zhou Daniel Hao$^{1,2}$%
\thanks{$^{1}$Authors are with DANiLab, University of Leicester, Leicester, UK
{\tt\small \{hz273,aamd2,d.hao\}@leicester.ac.uk}.}%
\thanks{$^{2}$Authors are with the School of Computing and Mathematical Sciences, University of Leicester, Leicester, UK.}%
\thanks{$^{3}$The author is with the School of Metallurgy and Materials, University of Birmingham, Birmingham, UK.
{\tt\small h.dong.1@bham.ac.uk}.}}

\begin{document}

\makeatletter
\maketitle

\begin{abstract}
Spoken instructions in robot-to-human handovers may specify either an object (``the cup'') or an intended use (``pour water'');
in both cases, successful handover requires the robot to infer the target object and the region remaining available for the human to hold.
If the robot grasps that hold region, the object could become awkward to receive and immediately use, potentially reducing perceived competence and trust;
if the gripper approaches too close to the receiving hand during delivery, perceived safety may also suffer.
We present \textbf{Intent-Handover}, which grounds unconstrained speech and visual scene context into explicit grasp and delivery constraints.
Given a spoken instruction and a scene observation, a vision--language model identifies the target object and the intended human-usage region.
A grasp optimization module then selects a feasible grasp keeping this region accessible while enforcing clearance from the predicted receiving hand.
During execution, the robot tracks upper-body key points to estimate the user's receiving pose and places the handover at an ergonomically feasible location.
In a within-subjects ablation study (n=30), human-usage region awareness increases perceived trust, hand-gripper collision avoidance increases perceived safety,
and interaction comfort is highest when both are enabled.
Website and code: \href{https://robot-future.github.io/intent-handover/}{https://robot-future.github.io/intent-handover/}.
\end{abstract}

\section{Introduction}

Robot-to-human (R2H) object handover is a fundamental capability in human-robot collaboration~\cite{ortenziObjectHandoversReview2021,duan2024human}.
Existing approaches attempt to build trustworthy handovers by improving grasp stability~\cite{9366406}, delivery safety~\cite{megyeri2025safety}, and user comfort~\cite{meng2024fast}, but have not fully considered human's post-handover usage intent.
Instructions such as ``pour water'' implicitly constrain the grasp region: the cup handle may be better left for the human to grasp.
A robot may grasp the object stably and deliver it successfully, but the human still needs to readjust their grasp posture before using it.
Prior studies suggest that such readjustment can reduce perceived competence and degrade trust~\cite{ortenzi2020grasp,scirobotics.aau9757}.
This raises a first concern: the robot may occupy the region the human intends to grasp.

A second concern arises during delivery: the robot end-effector may approach too close to the receiver's hand.
Insufficient clearance between the two can potentially reduce perceived safety and comfort~\cite{lasotaSurveyMethodsSafe2017,meng2024fast}.
These two concerns are separable: human-usage region awareness shapes perceived competence and trust, while hand-gripper collision avoidance shapes perceived safety and comfort.
Trustworthy R2H handover therefore requires satisfying both constraints simultaneously.

\begin{figure}[t]
    \centering
    \includegraphics[width=\linewidth]{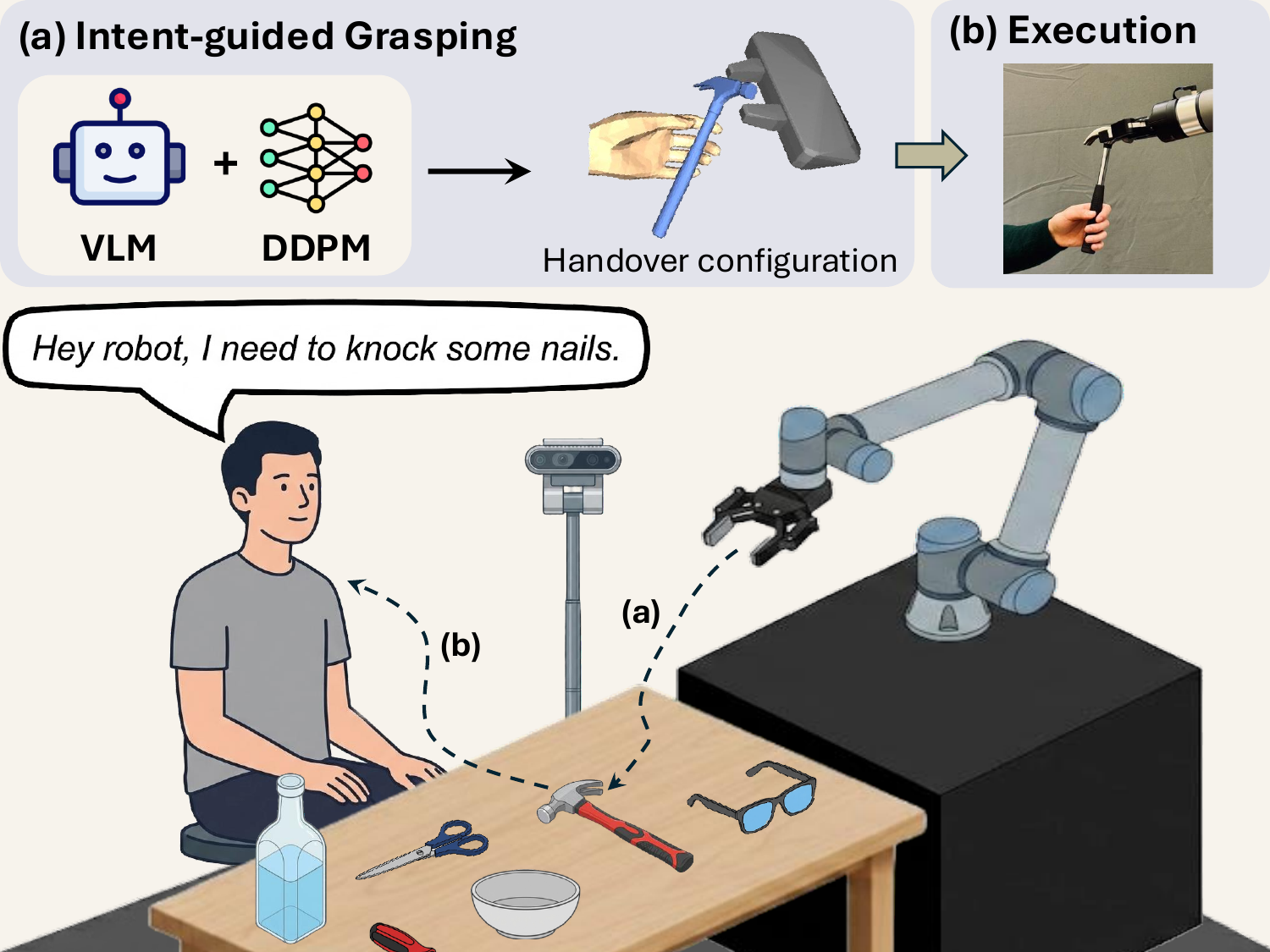}
    \caption{\textbf{Intent-Guided Robot-to-Human Object Handover.}
    (a) \textit{Intent-Guided Grasping}: A Vision--Language Model and diffusion model infer user intent and human grasp pose from speech and RGB inputs. 
    The system then optimizes the robot grasp configuration.
    (b) \textit{Execution}: The robot aligns the object with the receiving hand for ergonomic delivery.
    }
    \label{fig:system_overview}
\end{figure}

To address both concerns, the robot needs to consider which region the human intends to grasp and how the human is likely to grasp it.
Current handover systems can identify the target object from human verbal and visual cues~\cite{herreraLearningfindinggivingNaturalVisionspeechbased2024,langer2022let,fei2025large}, but may not reason about intended human grasp region.
Systems that account for grasp region typically rely on static affordance labels~\cite{lehotsky2023optimizing} or demonstration-derived contact maps~\cite{wang2024contacthandover}, which are difficult to infer from natural language.
In addition, requiring explicit object specification~\cite{herreraLearningfindinggivingNaturalVisionspeechbased2024} or auxiliary sensing such as wearable devices~\cite{zou2023novel,liuHumantorobotHandoversBased2025} may limit interaction naturalness.
Therefore, the core technical challenge remains: computing an optimal robot grasp from a spoken instruction and an RGB scene image that avoids the human-usage region and prevents hand-gripper collision.

In this paper, we present \textbf{Intent-Handover}, an R2H handover system that grounds spoken instructions into explicit spatial constraints and jointly optimizes for human-usage region awareness and hand-gripper collision avoidance.
As shown in Fig.~\ref{fig:system_overview}, the system operates in two phases.
In the intent-guided grasping phase (Sec.~\ref{subsec:intent_grasp}), a vision--language pipeline maps the spoken instruction and RGB scene image to a structured intent identifying the target object and its intended usage region.
A grasp optimization module then selects a grasp that avoids the inferred grasp region and minimizes proximity to the predicted receiving hand.
During execution, the robot estimates the user's receiving pose from skeletal key points and delivers the object at an ergonomically appropriate location.

We conduct a within-subjects ablation user study with 30 participants, independently disabling human-usage region awareness and hand-gripper collision avoidance, to isolate the effect of each constraint.
The study provides empirical evidence for two design hypotheses: 
(\textbf{H1}) users trust the robot more when it delivers the object in a grasp that is easy for them to use, 
and (\textbf{H2}) users feel less safe and less comfortable when the robot does not account for hand-gripper collision avoidance during delivery.

The contributions of this work are as follows.
\begin{enumerate}
\item
\textbf{Intent grounding.}
A pipeline from speech and RGB scene image that predicts the human's natural grasp pose for the intended object, jointly resolving the target object and its usage region from unconstrained spoken instructions.
\item
\textbf{Grasp optimization from inferred intent.}
A grasp selection method that enforces human-usage region awareness and hand-gripper collision avoidance, with both constraints derived from natural language rather than predefined labels or demonstrations.
\item
\textbf{Empirically validated design principles.}
A within-subjects ablation user study demonstrating that (1) human-usage region awareness enhances perceived trust, (2) hand-gripper collision avoidance enhances perceived safety, and (3) comfortable handover requires both constraints simultaneously.
\end{enumerate}
\section{Related Work}
\textbf{Robot-to-Human Object Handovers.}
R2H handover requires the robot to infer the receiver's intent and deliver the object safely, comfortably, and in a functionally appropriate manner \cite{duan2024human, ortenzi2020grasp}, anticipating the shared grasp pose, usage intent, and handover timing \cite{ortenziObjectHandoversReview2021}.
Intent has been inferred from language \cite{tulbure2025llm}, gestures \cite{songPredictingHumanIntention2013}, or eye gaze \cite{liInferenceManipulationIntent2020};
compared to physical modalities, natural language conveys intent at a higher level of semantic abstraction, allowing users to express intended uses directly.
In this work, we infer handover intent from language and visual scene context using vision--language models (VLMs), jointly enforcing human-usage region awareness and hand-gripper collision avoidance.

\textbf{Handover Intent Identification.}
Human receivers typically communicate handover intent by extending their hands toward the robot \cite{strabala2013towards,strabala2012learning,roy2020investigating};
some systems instead infer intent from verbal instructions using speech recognition to identify the target object \cite{herreraLearningfindinggivingNaturalVisionspeechbased2024, langer2022let}.
These systems typically rely on external sensing devices~\cite{zou2023novel,liuHumantorobotHandoversBased2025} and require users to name the target object explicitly, without inferring the intended usage region.
Vision--language models (VLMs) infer object identity and usage intent without auxiliary hardware.
Recent approaches ground handover intent via structured prompting~\cite{tulbure2025llm}, conversational reasoning~\cite{lakhnati2024exploring}, or predefined skills~\cite{fei2025large}.

However, they identify the target object only, and do not infer which part of the object the user intends to grasp.
We introduce a hierarchical prompting strategy that produces structured intent outputs encoding the target object and the region the user will grasp, providing a language prior for predicting the human's natural grasp pose to constrain robot grasp selection.

\textbf{Grasp Pose Prediction.}
Most R2H handover systems predict a shared grasp pose to guide handover execution \cite{megyeri2025safety}.
Meng et al. \cite{meng2022fast, meng2024fast} predict a receiving hand pose using GANHand \cite{coronaGanhandPredictingHuman2020} and select a grasp opposite the hand direction for fast delivery, but do not consider object affordance.
Lehotsky et al. \cite{lehotsky2023optimizing} target large stable regions identified by AffNet-DR \cite{christensenLearningSegmentObject2022} to improve grasp usability, but rely on static labels predefined per object.
ContactHandover \cite{wang2024contacthandover} reranks candidate grasps via a hand--object contact map to avoid unintended hand contact, but does not consider which region the user needs for the intended task.
None of these methods jointly enforces human-usage region awareness and hand-gripper collision avoidance as constraints derived from the user's spoken intent.
We derive both constraints from natural language rather than from predefined labels or observed demonstrations.
\begin{figure*}[t]
    \centering
    \includegraphics[width=\linewidth]{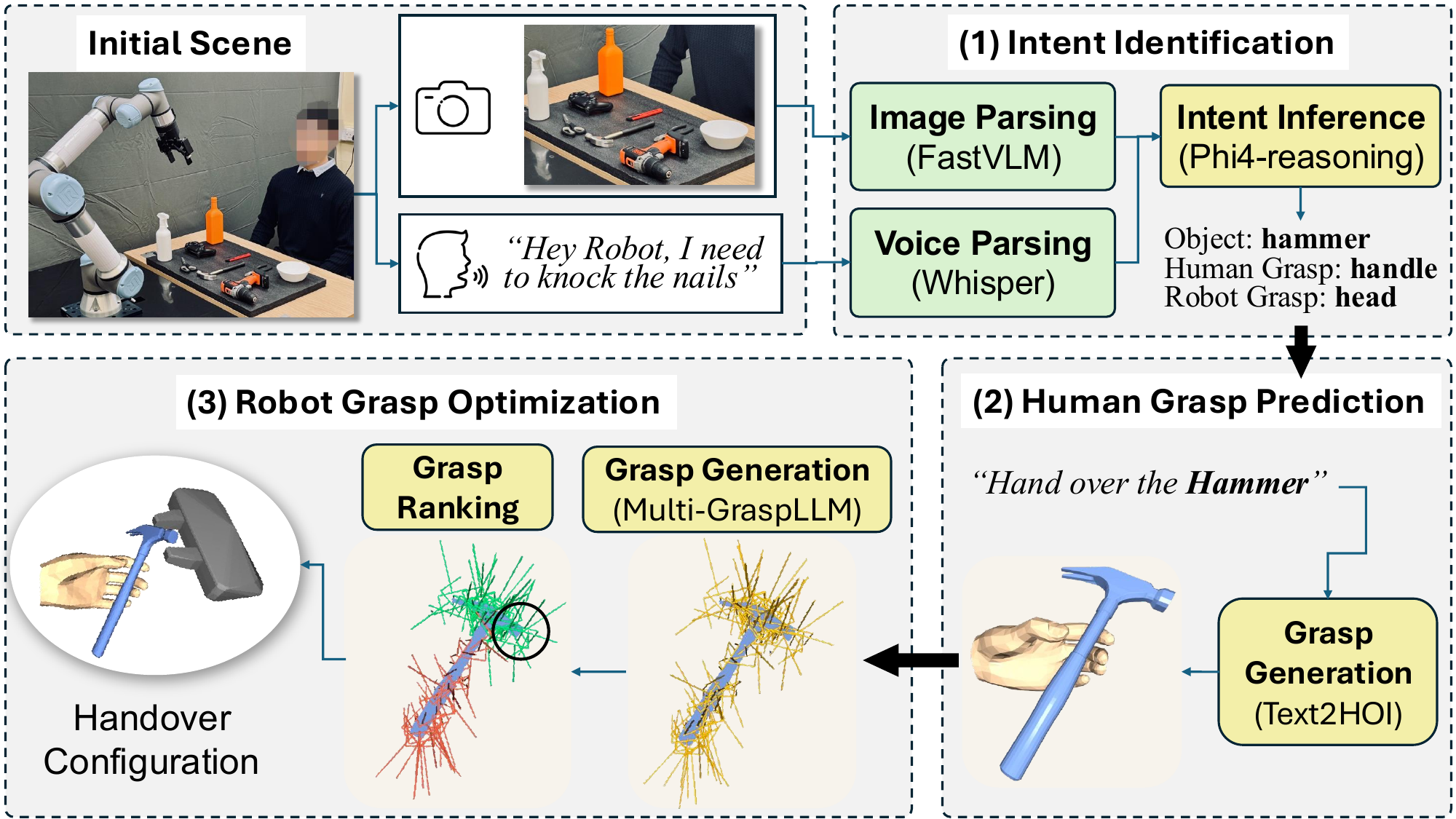}
    \caption{
    \textbf{Intent-Guided Grasping Phase.}
    Given a natural language instruction and a scene image, the system proceeds in three steps.
    (1) A vision-language model (VLM, Phi4-Reasoning) identifies the target object and infers the intended usage region from the instruction and scene image.
    (2) A diffusion model (Text2HOI) conditioned on language predicts the expected pose of the receiving hand.
    (3) Candidate robot grasp poses are evaluated based on the predicted hand pose and the intended usage region; the optimal grasp is selected for execution.
    }
    \label{fig:grasping_phase}
\end{figure*}

\section{Methodology}
The Intent-Handover consists of two key stages: 
intent-guided grasping phase and execution phase, as shown in Fig.~\ref{fig:system_overview}. 
During the intent-guided grasping phase, the system first translates user instructions and scene images into textual handover intent, 
then predicts the corresponding receiving hand pose, and optimizes a robotic grasp.
During the execution phase, the system first grasps the object, 
then localizes the receiver's hand and aligns it with the predicted pose, ultimately completing the handover.

\subsection{\textbf{Intent-Guided Grasping Phase}}
\label{subsec:intent_grasp}

The intent-guided grasping phase aims to generate a grasp configuration that aligns with the receiver's intended use of the object while minimizing interference with the human hand.
As illustrated in Fig.~\ref{fig:grasping_phase}, given a spoken instruction and an RGB observation of the tabletop scene, 
the system first infers the textual handover intent, then predicts a corresponding receiver hand pose together with a set of candidate robot grasps.
A final score is computed for each candidate grasp by combining human-usage region awareness and hand-gripper collision avoidance criteria.
The grasp with the highest score is selected for execution.

\subsubsection{\textbf{Intent Identification}}

We translate multimodal user input into structured handover intent using a hierarchical prompting framework.

The process begins when Whisper~\cite{radford2023whisper} transcribes the spoken instruction into text $T$. 
An Intel RealSense Depth Camera D435i captures the scene image $I \in \mathbb{R}^{H \times W \times C}$. 
FastVLM~\cite{fastvlm:vasu2025fastvlmefficientvisionencoding} parses the scene image $I$ to produce a visual description of the objects.
The text-image pair $(T, I)$ is then processed by Phi4-Reasoning~\cite{phi4reasoning:abdin2025phi4reasoningtechnicalreport},
which refines the raw instruction into a structured intent representation $T'$.

The processed intent $T'$ encodes the predicted object category $\hat{o}$ and a textual usage representation $\hat{u}$
that specifies which object regions are compatible with the intended use, following a fixed template filled by the VLM (e.g., ``\textit{I want to drill a hole}'' $\to$ ``\textit{the robot grasps the \textbf{drill head}, the human grasps the \textbf{handle}}'').

\subsubsection{\textbf{Human Grasp Prediction}}

To predict the human receiving pose, we train Text2HOI~\cite{cha2024text2hoi}, a language-conditioned diffusion model. 
Given structured intent $T'$ and object point cloud $\mathbf{x}_{obj}$, it generates a plausible interaction using a transformer-based denoiser conditioned on language embeddings and a geometric contact prior.

The output of the model is the predicted MANO parameter vector $\mathbf{x}_{hand}$ of the receiving hand, which defines its global pose and articulated joint configuration in 3D space. 
We train the interaction generation components of Text2HOI on a handover dataset constructed by integrating data from H2O~\cite{Kwon_2021_ICCV}, GRAB~\cite{taheriGRABDatasetWholebody2020}, and ARCTIC~\cite{fan2023arctic} .
To construct the training dataset, we manually curated 400 grasp-related trajectories from these datasets, retaining only frames where the hand is in contact with the object and discarding approach and post-use frames.
Each retained sequence was manually annotated with a structured intent prompt $T'$.

During training, the model learns to reconstruct $\mathbf{x}_{hand}$ from $(T', \mathbf{x}_{obj})$, capturing the distribution of natural receiving poses.
We use $\mathbf{T}_{ab} \in SE(3)$ to denote the rigid transform that maps coordinates from frame $b$ to frame $a$.
Finally, the predicted pose is transformed to the robot base frame via 
$\mathbf{T}_{rh} = \mathbf{T}_{rw}\mathbf{T}_{wo}\mathbf{T}_{oh}$, 
where $\mathbf{T}_{oh}$, $\mathbf{T}_{wo}$, and $\mathbf{T}_{rw}$ denote the hand-to-object, object-to-world, and world-to-robot transforms, respectively.

\subsubsection{\textbf{Robot Grasp Optimization}}

For each object $o$ in our dataset, we use the grasp annotations provided by Multi-GraspLLM~\cite{liMultiGraspLLMMultimodalLLM2024} as the candidate grasp set 
$\mathcal{G}_{o} = \{ g_i \}_{i=1}^{N}$, where each $g_i$ is a 6-DoF gripper pose defined in the object coordinate frame. 
Let $\mathcal{S}_o$ denote the object surface point set and let $\mathcal{R}_o \subset \mathcal{S}_o$ denote the human-usage region specified by the structured intent $T'$.

\textbf{Human-usage Region Awareness.}
For each candidate grasp $g_i$, we verify geometric feasibility and ensure the grasp lies outside the human-usage region. 
Let $w(g_i)$ denote the object width measured along the gripper closing direction under $g_i$, 
and let $\mathbf{x}_{int}(g_i) \in \mathcal{S}_o$ denote the object surface point intersected by the gripper approach axis. 
A grasp $g_i$ is considered valid if it satisfies
\[
w(g_i) \le w_{\max}
\;\land\;
\mathbf{x}_{int}(g_i) \notin \mathcal{R}_o,
\]
where $w_{\max}$ is the maximum gripper aperture.

\textbf{Hand-Gripper Collision Avoidance.}
Let $\mathbf{v}_h \in \mathbb{R}^3$ denote the wrist-to-middle-finger direction of the predicted receiving hand,
and let $\mathbf{v}_{g_i} \in \mathbb{R}^3$ denote the approach direction of $g_i$.
Let $\mathbf{p}_h$ and $\mathbf{p}_{g_i}$ denote the centers of the hand and gripper, respectively.
We define the avoidance cost
\[
A(g_i)
=
\frac{\mathbf{v}_{g_i} \cdot \mathbf{v}_h}{\|\mathbf{v}_{g_i}\| \|\mathbf{v}_h\|}
- \|\mathbf{p}_{g_i} - \mathbf{p}_h\|.
\]
The first term is minimized when the gripper and hand approach from opposite directions ($\cos\theta \to -1$), maximizing spatial separation during handover.
The second term favors grasps whose center is far from the predicted hand center.

\textbf{Final selection.}
The optimal grasp minimizes the avoidance cost among all feasible candidates:
\[
g^{\star}
=
\arg\min_{g_i \in \mathcal{G}_{o}}
A(g_i)
\quad
\text{s.t.}
\quad
w(g_i) \le w_{\max},
\;
\mathbf{x}_{int}(g_i) \notin \mathcal{R}_o.
\]
The selected grasp $g^{\star}$ is passed to the execution module.

\begin{figure}[t]
    \centering
    \includegraphics[width=\linewidth]{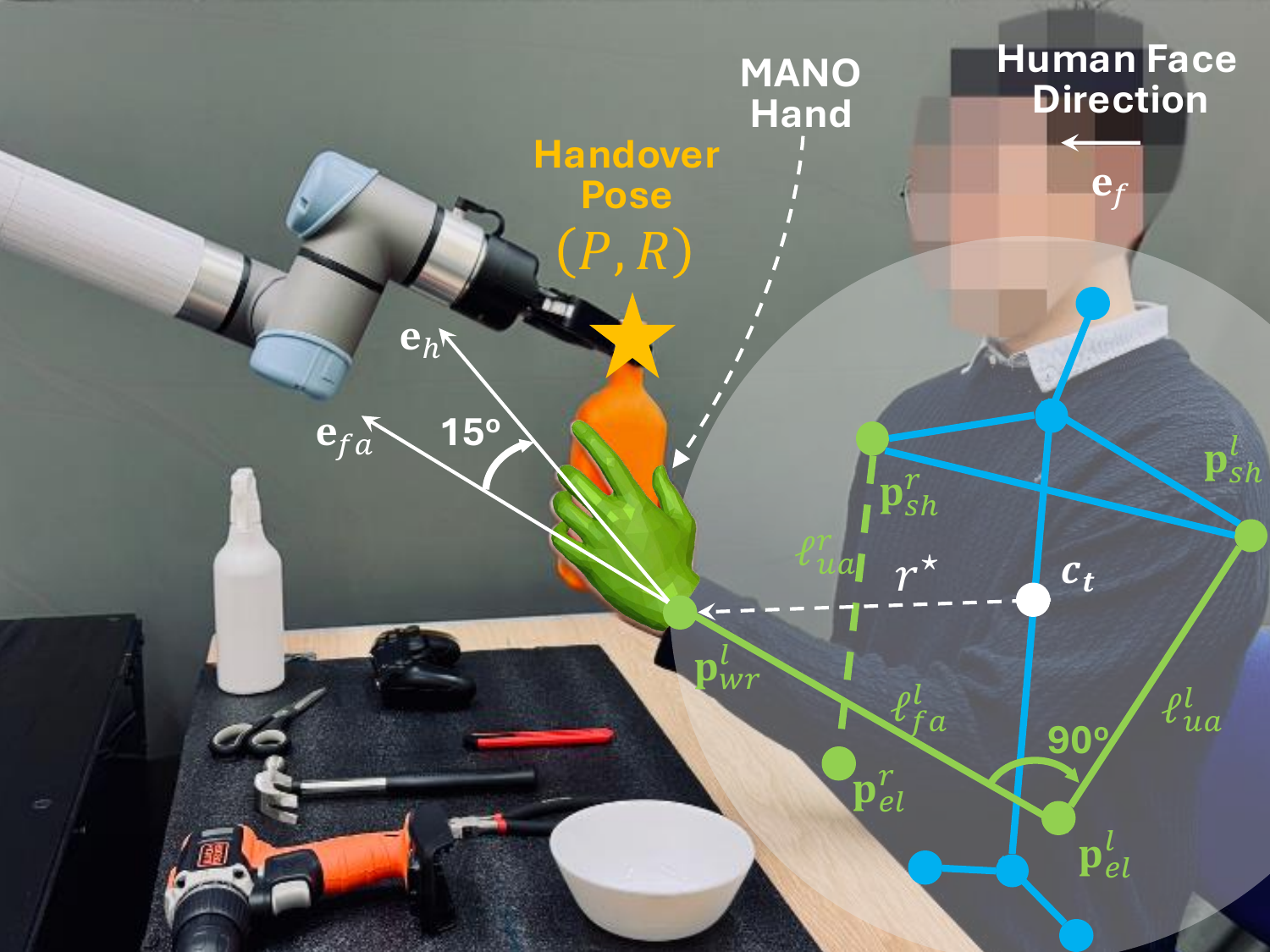}
    \caption{\textbf{Execution Phase: Handover Pose Estimation.}
    The handover position $P$ and orientation $R$ are derived from skeletal keypoints to ensure comfortable reachability. See Sec.~\ref{subsec:execution_phase} for details.
    }
    \label{fig:execution_phase}
\end{figure}

\begin{figure*}[t]
    \centering
    \includegraphics[width=\textwidth]{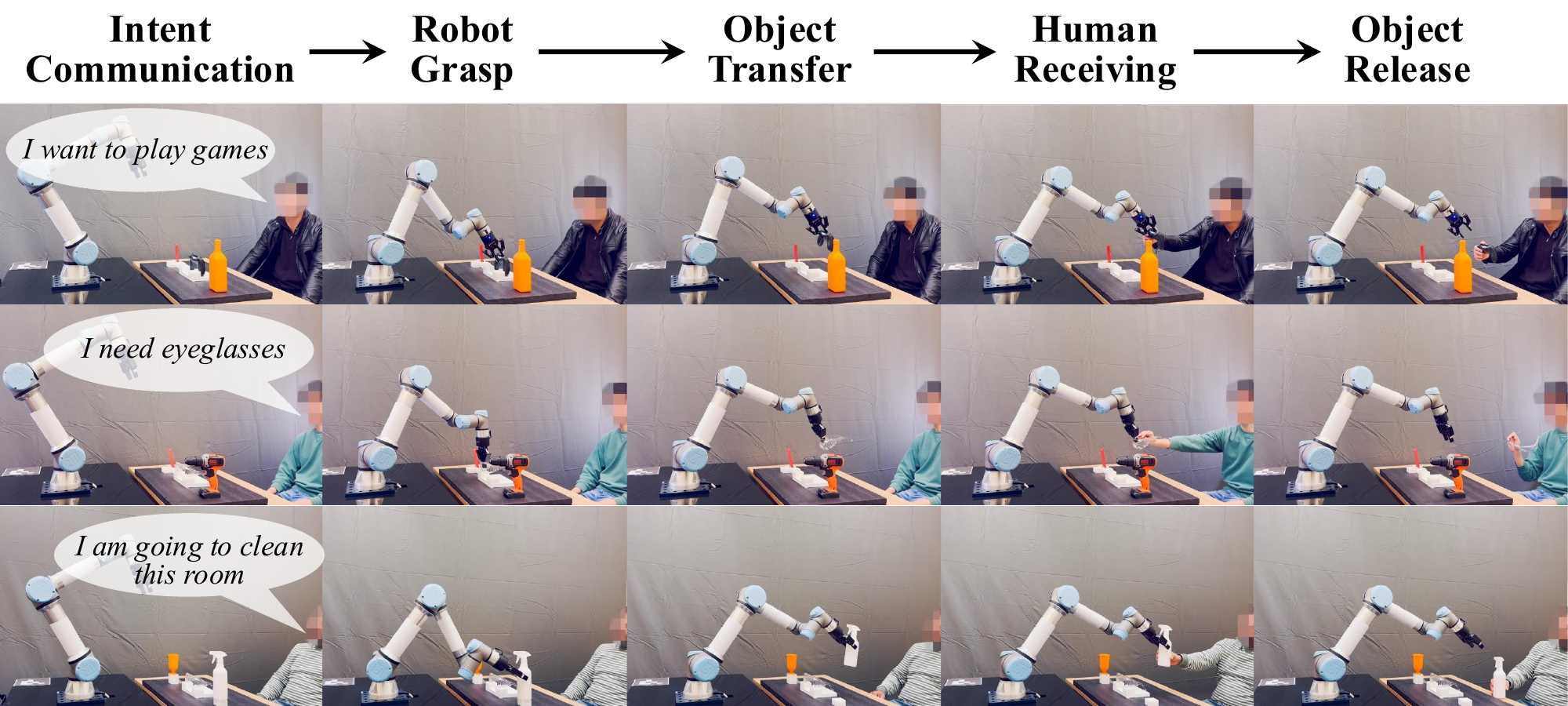}
    \caption{
    \textbf{User Study Procedure.}
    Each trial follows five stages: intent communication, robot grasp, object transfer, human receiving, and object release.
    Each row shows a complete handover sequence for one participant.
    Three representative participants are shown.
    }
    \label{fig:procedure_stages}
\end{figure*}

\subsection{\textbf{Execution Phase}}
\label{subsec:execution_phase}

As illustrated in Fig.~\ref{fig:execution_phase}, the robot estimates the handover position and orientation using an RGB perception setup with depth estimation.
An Intel RealSense D435i camera is mounted at a height of $2.30\,\mathrm{m}$ above the floor and positioned $1.90\,\mathrm{m}$ away from the calibrated world-frame origin in the horizontal plane, 
at a $45^\circ$ diagonal offset from the table front edge.
Its optical axis is oriented toward the center of the workspace at a $45^\circ$ horizontal angle.
All observations are transformed into a unified world frame via extrinsic calibration, 
and 3D skeletal key points are extracted using MediaPipe \cite{lugaresi2019mediapipe}.

\subsubsection{\textbf{Handover Position}}
\label{subsubsec:handover_position}

The handover position is defined to ensure reachability and arm comfort for a seated user.
Following ergonomic principles \cite{katayama2003optimization} and prior handover studies \cite{wang2024contacthandover},
we model the human reachable workspace as a sphere \cite{liu2017seated,hsiao1991evaluating}
centered at an upper-body reference frame.

Let $\mathbf{p}_{sh}^l, \mathbf{p}_{sh}^r \in \mathbb{R}^3$ denote the left and right shoulder positions.
The shoulder midpoint is
\[
\mathbf{c}_{sh} = \frac{\mathbf{p}_{sh}^l + \mathbf{p}_{sh}^r}{2}.
\]
Let $z_{\text{desk}}$ denote the desk height.
To account for seated tabletop interaction, the torso reference center is defined as
\[
\mathbf{c}_t =
\begin{bmatrix}
(\mathbf{c}_{sh})_x,
(\mathbf{c}_{sh})_y,
\frac{(\mathbf{c}_{sh})_z + z_{\text{desk}}}{2}
\end{bmatrix}.
\]

Let $\mathbf{p}_{el}, \mathbf{p}_{wr}$ denote the elbow and wrist positions,
and define the upper-arm and forearm lengths as
\[
\ell_{ua} = \|\mathbf{p}_{el} - \mathbf{c}_{sh}\|, 
\quad
\ell_{fa} = \|\mathbf{p}_{wr} - \mathbf{p}_{el}\|.
\]

To avoid extreme extension, we select a mid-range posture with an elbow flexion of approximately $90^\circ$,
which is regarded as a neutral configuration in ergonomics \cite{baker2013relationship}.
The corresponding comfortable reach radius is
\[
r^\star = \sqrt{\ell_{ua}^2 + \ell_{fa}^2}.
\]

Let $\mathbf{e}_f$ denote the unit horizontal facing direction of the human, computed as the horizontal component of the vector perpendicular to the left-right shoulder axis $(\mathbf{p}_{sh}^r - \mathbf{p}_{sh}^l)$, pointing forward.
The delivery position is defined as
\[
P = \mathbf{c}_t + r^\star \mathbf{e}_f.
\]

The predicted handover configuration specifies a fixed object-to-end-effector transform $\mathbf{T}_{eo}$.
Therefore, the delivery position $P$ uniquely determines the target end-effector position.

\subsubsection{\textbf{Handover Orientation}}
\label{subsubsec:handover_orientation}

The user is assumed to receive the object at delivery position $P$.
The hand orientation is determined from the forearm direction estimated from upper-limb keypoints.

Let $\mathbf{e}_{fa}$ denote the unit forearm direction from elbow to wrist.
Natural grasping involves a slight wrist extension rather than strict collinearity with the forearm axis; we adopt $15^\circ$ as it minimizes musculoskeletal cost while remaining within the comfortable wrist extension range \cite{poppEffectsWristPosture2023},
\[
\mathbf{e}_h = R_{ext}(15^\circ)\,\mathbf{e}_{fa},
\]
where $R_{ext}(15^\circ)$ denotes a rotation of $15^\circ$ in the wrist flexion–extension plane.

Let $\mathbf{e}_{mano}$ denote the canonical wrist-to-middle-finger axis of the MANO model.
The target hand orientation $R \in SO(3)$ is the minimum-angle rotation that aligns $\mathbf{e}_{mano}$ with $\mathbf{e}_h$:
\[
R\,\mathbf{e}_{mano} = \mathbf{e}_h, \quad
R = \arg\min_{\tilde{R}} \|\tilde{R} - I\|_F
\;\;\text{s.t.}\;\; \tilde{R}\,\mathbf{e}_{mano} = \mathbf{e}_h.
\]

The predicted handover configuration specifies a fixed hand-to-end-effector transform $\mathbf{T}_{eh}$.
Therefore, the target end-effector orientation is directly determined by $R$.

\section{USER STUDY}

All participants provided written informed consent prior to participation.

\subsection{Research Questions and Hypotheses}
We isolate the effects of two grasp optimization strategies (human-usage region awareness and hand-gripper collision avoidance) through two research questions.

(\textbf{RQ1}) To what extent does the robot's grasp region affect user trust during handover?
Correctly inferring the intended object alone is insufficient; grasping an inappropriate region may still undermine trust.
\begin{itemize}
    \item \textbf{H1.}
    Users trust the robot more when it delivers the object in a grasp that is easy for them to use.
\end{itemize}

(\textbf{RQ2}) To what extent does failing to consider the user's hand affect perceived comfort and safety?
Even if users can adjust their receiving pose, neglecting hand-gripper collision avoidance may reduce psychological safety.
\begin{itemize}
    \item \textbf{H2.}
    Users feel less safe and comfortable when the robot does not optimize for hand-gripper collision avoidance.
\end{itemize}

\begin{figure*}[!t]
\centering
\definecolor{fscol}{RGB}{232,141,179}
\definecolor{a1col}{RGB}{126,192,144}
\definecolor{a2col}{RGB}{121,170,229}
\definecolor{a3col}{RGB}{184,153,221}

\begin{minipage}[t]{0.495\textwidth}
\centering
\begin{tikzpicture}
\begin{axis}[
    ybar,
    bar width=12pt,
    width=\linewidth,
    height=6.0cm,
    ymin=0, ymax=8.6,
    ytick={0,1,2,3,4,5,6,7},
    ylabel={Likert Score (1--7)},
    symbolic x coords={Competence,Reliability,Benevolence},
    xtick=data,
    xticklabel style={font=\footnotesize,align=center,text height=1.5ex,text depth=.25ex},
    title={(a) Trust-Related Dimensions (RQ1)},
    title style={font=\small},
    ymajorgrids=true,
    major grid style={gray!25,line width=0.3pt},
    axis line style={black!60},
    tick style={black!60},
    enlarge x limits=0.24,
    axis background/.style={fill=gray!6},
    legend style={at={(0.02,0.02)},anchor=south west,draw=gray!60,fill=white,font=\scriptsize},
    legend cell align={left},
    legend image code/.code={
        \draw[#1,draw=black!55,line width=0.35pt] (0cm,-0.08cm) rectangle (0.16cm,0.08cm);
    },
    clip=false
]
\addplot+[bar shift=-12pt,draw=black!55,line width=0.35pt,fill=fscol,error bars/.cd,y dir=both,y explicit] coordinates {
    (Competence,5.8) +- (0,0.60)
    (Reliability,5.6) +- (0,0.95)
    (Benevolence,5.3) +- (0,1.25)
};
\addplot+[bar shift=0pt,draw=black!55,line width=0.35pt,fill=a1col,error bars/.cd,y dir=both,y explicit] coordinates {
    (Competence,5.0) +- (0,0.80)
    (Reliability,5.3) +- (0,0.75)
    (Benevolence,5.2) +- (0,1.10)
};
\addplot+[bar shift=12pt,draw=black!55,line width=0.35pt,fill=a3col,error bars/.cd,y dir=both,y explicit] coordinates {
    (Competence,4.6) +- (0,1.45)
    (Reliability,4.7) +- (0,0.85)
    (Benevolence,5.0) +- (0,1.35)
};
\legend{FS,A1,A3}

\node[font=\footnotesize\bfseries,draw=gray!60,fill=white,rounded corners=1pt,inner xsep=1.8pt,inner ysep=2.4pt,minimum height=7.5mm,align=center,anchor=north] at (axis cs:Competence,8.20) {A1:***\\A3:***};
\node[font=\footnotesize\bfseries,draw=gray!60,fill=white,rounded corners=1pt,inner xsep=1.8pt,inner ysep=2.4pt,minimum height=7.5mm,align=center,anchor=north] at (axis cs:Reliability,8.20) {A1:*\\A3:***};
\node[font=\footnotesize\bfseries,draw=gray!60,fill=white,rounded corners=1pt,inner xsep=1.8pt,inner ysep=2.4pt,minimum height=7.5mm,align=center,anchor=north] at (axis cs:Benevolence,8.20) {A1:n.s.\\A3:n.s.};
\end{axis}
\end{tikzpicture}
\end{minipage}
\hfill
\begin{minipage}[t]{0.495\textwidth}
\centering
\begin{tikzpicture}
\begin{axis}[
    ybar,
    bar width=12pt,
    width=\linewidth,
    height=6.0cm,
    ymin=0, ymax=8.6,
    ytick={0,1,2,3,4,5,6,7},
    xmin=0, xmax=3,
    xtick={1,2},
    xticklabels={Safety,Comfort},
    xticklabel style={font=\footnotesize,align=center,text height=1.5ex,text depth=.25ex},
    title={(b) Safety and Comfort (RQ2)},
    title style={font=\small},
    ymajorgrids=true,
    major grid style={gray!25,line width=0.3pt},
    axis line style={black!60},
    tick style={black!60},
    enlarge x limits=false,
    axis background/.style={fill=gray!6},
    legend style={at={(0.02,0.02)},anchor=south west,draw=gray!60,fill=white,font=\scriptsize},
    legend cell align={left},
    legend image code/.code={
        \draw[#1,draw=black!55,line width=0.35pt] (0cm,-0.08cm) rectangle (0.16cm,0.08cm);
    },
    clip=false
]
\addplot+[bar shift=-12pt,draw=black!55,line width=0.35pt,fill=fscol,error bars/.cd,y dir=both,y explicit] coordinates {
    (1,6.1) +- (0,0.45)
    (2,5.6) +- (0,1.10)
};
\addplot+[bar shift=0pt,draw=black!55,line width=0.35pt,fill=a2col,error bars/.cd,y dir=both,y explicit] coordinates {
    (1,4.7) +- (0,1.35)
    (2,5.4) +- (0,0.85)
};
\addplot+[bar shift=12pt,draw=black!55,line width=0.35pt,fill=a3col,error bars/.cd,y dir=both,y explicit] coordinates {
    (1,3.9) +- (0,1.60)
    (2,4.8) +- (0,1.05)
};
\legend{FS,A2,A3}

\node[font=\footnotesize\bfseries,draw=gray!60,fill=white,rounded corners=1pt,inner xsep=1.8pt,inner ysep=2.4pt,minimum height=7.5mm,align=center,anchor=north] at (axis cs:1,8.20) {A2:***\\A3:***};
\node[font=\footnotesize\bfseries,draw=gray!60,fill=white,rounded corners=1pt,inner xsep=1.8pt,inner ysep=2.4pt,minimum height=7.5mm,align=center,anchor=north] at (axis cs:2,8.20) {A2:n.s.\\A3:**};
\end{axis}
\end{tikzpicture}
\end{minipage}

\caption{\textbf{User-Study Results: Trust, Perceived Safety, and Interaction Comfort.}
(a) Mean trust ratings (competence, reliability, benevolence) across FS, A1, and A3.
(b) Mean perceived safety and interaction comfort ratings across FS, A2, and A3.
Bars show mean Likert scores (1--7); error bars show $\pm$SD.
Annotation boxes show Bonferroni-corrected post-hoc comparisons against FS ($^{*}p<.05$, $^{**}p<.01$, $^{***}p<.001$, n.s. not significant; all $p$-values are Bonferroni-adjusted).
See TABLE~\ref{tab:pairwise_comparison} for full statistics.}
\vspace{-5pt}
\label{fig:results_barchart}
\end{figure*}


\subsection{Procedure}
To evaluate the impact of our system on users' perceived trust and safety during physical handovers, we conducted a mixed-design user study.
All participants experienced four system variants: the \textbf{Full Strategy} (\textbf{FS}) and three ablations:
\begin{itemize}
    \item \textbf{Ablation 1 (A1): No Human-usage Region Awareness:} 
    The robot does not consider whether a predicted grasp supports easy human use after handover.
    For example, it may grasp the blade of scissors rather than the handle, hindering user reception and use.
    \item \textbf{Ablation 2 (A2): No Hand-Gripper Collision Avoidance:} 
    The robot disables constraints for hand-gripper collision avoidance, and the predicted grasp may result in contact with the user's hand.
    \item \textbf{Ablation 3 (A3): A1 + A2:}
    The robot selects a grasp solely based on whether it can grasp the object, ignoring both constraints.
\end{itemize}

The hardware platform consists of a Universal Robots UR5e arm equipped with a Robotiq 2F-85 parallel-jaw gripper.
The object dataset comprises 16 everyday objects spanning tools, kitchenware, and household items.
Object selection was randomized across trials such that every object appeared under every condition at least once across the full participant pool.
In the experimental setup, three objects from this dataset are placed adjacently in front of the user with specific spatial intervals.
Fig.~\ref{fig:procedure_stages} illustrates the five stages of a complete handover trial: intent communication, robot grasp, object transfer, human receiving, and object release.
Some items, such as game controllers and knives, are elevated or supported, facilitating smooth grasp execution.
Participants then express their intent in natural language by either naming the desired object or describing how they plan to use it (e.g., "\textit{I want scissors}" or "\textit{I want to cut paper"}).
The system generates a corresponding grasp configuration, actuates the robotic arm to grasp the object from a fixed initial position, and delivers it to the human receiver.
An embedded force-torque sensor within the end effector dictates the object release timing by triggering the gripper to release when the sensed force exceeds a preset threshold.

To mitigate order effects, the presentation order of the four conditions was independently randomized for each participant.
Each participant completed 20 trials (five for each condition). After each block of five trials for a given condition, they completed a questionnaire.

\begin{table}[!t]
\centering
\caption{Pairwise comparison results for the user-study conditions (7-point Likert means; $n=30$, $\mathrm{df}=29$; Bonferroni-corrected post-hoc comparisons following significant repeated-measures ANOVAs; n.s. not significant). Trust dimensions (competence, reliability, benevolence) are from MDMT v2; perceived safety and interaction comfort are from separate validated scales.}
\label{tab:pairwise_comparison}
\setlength{\tabcolsep}{1.0pt}
\renewcommand{\arraystretch}{1.22}
\footnotesize
\begin{tabular*}{\columnwidth}{@{\extracolsep{\fill}}>{\raggedright\arraybackslash}p{0.27\columnwidth}c@{\hspace{2pt}}c>{\centering\arraybackslash}p{0.10\columnwidth}>{\centering\arraybackslash}p{0.10\columnwidth}>{\centering\arraybackslash}p{0.10\columnwidth}c@{}}
\toprule
\textbf{Measure} & \textbf{FS (M)} & \textbf{Abl. (M)} & \textbf{$t$} & \textbf{$p$} & \textbf{\mbox{$d_z$}} & \textbf{Effect} \\
\midrule
\multicolumn{7}{@{}l@{}}{\colorbox{gray!20}{\parbox{\dimexpr\columnwidth-2\fboxsep-2\tabcolsep\relax}{\textbf{MDMT (Trust, FS vs A1)}}}}\\
Competence & $5.8$ & $5.0$ & 4.20 & $<.001$ & 0.77 & FS $>$ A1 \\
Reliability & $5.6$ & $5.3$ & 2.45 & $.02$ & 0.45 & FS $>$ A1 \\
Benevolence & $5.3$ & $5.2$ & 0.60 & $.55$ & 0.11 & n.s. \\
\midrule
\multicolumn{7}{@{}l@{}}{\colorbox{gray!20}{\parbox{\dimexpr\columnwidth-2\fboxsep-2\tabcolsep\relax}{\textbf{MDMT (Trust, FS vs A3)}}}}\\
Competence & $5.8$ & $4.6$ & 6.00 & $<.001$ & 1.10 & FS $>$ A3 \\
Reliability & $5.6$ & $4.7$ & 5.20 & $<.001$ & 0.95 & FS $>$ A3 \\
Benevolence & $5.3$ & $5.0$ & 2.10 & $.04$ & 0.38 & n.s. \\
\midrule
\multicolumn{7}{@{}l@{}}{\colorbox{gray!20}{\parbox{\dimexpr\columnwidth-2\fboxsep-2\tabcolsep\relax}{\textbf{Safety \& Comfort (FS vs A2)}}}}\\
Perceived Safety & $6.1$ & $4.7$ & 6.58 & $<.001$ & 1.20 & FS $>$ A2 \\
Interaction~Comfort & $5.6$ & $5.4$ & 1.45 & $.16$ & 0.26 & n.s. \\
\midrule
\multicolumn{7}{@{}l@{}}{\colorbox{gray!20}{\parbox{\dimexpr\columnwidth-2\fboxsep-2\tabcolsep\relax}{\textbf{Safety \& Comfort (FS vs A3)}}}}\\
Perceived Safety & $6.1$ & $3.9$ & 7.67 & $<.001$ & 1.40 & FS $>$ A3 \\
Interaction~Comfort & $5.6$ & $4.8$ & 3.65 & $<.01$ & 0.67 & FS $>$ A3 \\
\bottomrule
\end{tabular*}
\end{table}

\subsection{Evaluation Measures}
We assessed participants' trust, perceived safety, and interaction comfort using validated questionnaire instruments.
After completing each system condition (FS, A1, A2, A3), participants responded to a questionnaire battery covering all measures, with each item rated on a 7-point Likert scale (1 = strongly disagree, 7 = strongly agree).
Shapiro-Wilk tests indicated no significant deviation from normality for any measure in any condition ($p > .05$).
For each measure, we first conducted a one-way repeated-measures ANOVA across the four conditions; where a significant main effect was found, we performed Bonferroni-corrected pairwise comparisons.

To assess \textbf{RQ1} (trust), we employed the Multi-Dimensional Measure of Trust (MDMT) v2 \cite{malle2021multidimensional}, selecting three subscales: \textit{competence} (perceived capability of grasping and delivering appropriately), \textit{reliability} (perceived consistency across trials), and \textit{benevolence} (perceived intent to act in the user's interest).

To assess \textbf{RQ2}, we measured \textit{perceived safety} \cite{lasotaSurveyMethodsSafe2017} (how safe participants felt regarding hand-gripper contact) and \textit{interaction comfort} \cite{strabala2013towards} (how natural and effortless it was to receive and immediately use the object).

\section{RESULTS}
\label{sec:results}

A total of 30 participants were recruited to participate without compensation through local recruitment.
Participants self-identified as men (n=17) and women (n=13).
Ages of the participants ranged from 22 to 35 years (M = 26.3, SD = 2.5).
For each measure, we first ran a one-way repeated-measures ANOVA across the four conditions; where a significant main effect was found ($p<.05$), we followed up with Bonferroni-corrected pairwise comparisons.

\subsection{Effect of Grasp Region on Trust (RQ1)}
\label{subsec:results_rq1}
To answer RQ1, we compared the full strategy (\textbf{FS}) against \textbf{A1} and \textbf{A3} using Bonferroni-adjusted pairwise tests (TABLE~\ref{tab:pairwise_comparison}). In the \textbf{FS} vs \textbf{A1} contrast, dropping human-usage region awareness lowers perceived \emph{competence} ($p<.001$) and \emph{reliability} ($p=.02$) but leaves \emph{benevolence} unchanged ($p=.55$), meaning participants felt the robot was less capable and dependable, while the impression of considerateness stayed the same. 
In the \textbf{FS} vs \textbf{A3} contrast, the drops in \emph{competence} ($p<.001$) and \emph{reliability} ($p<.001$) are larger, while \emph{benevolence} did not reach significance after Bonferroni correction ($p=.04 > .025$), meaning capability and dependability fall further when both strategies are absent, while perceived care remains unaffected.
Overall, \textit{competence} is the most sensitive trust dimension; \textit{benevolence} did not reach significance in either contrast after Bonferroni correction, though the FS vs A3 comparison ($p=.04$) suggests a possible trend that warrants further investigation. These results support \textbf{H1}.
This is likely because participants treat the grasp region as a signal of ``does the robot understand how I will use this,'' so \textit{competence} and \textit{reliability} respond to task-level failures, while the \textit{benevolence} judgment remains stable regardless of the ablation.
Results are visualized in Fig.~\ref{fig:results_barchart}.

\begin{figure}[t]
    \centering
    \includegraphics[width=\linewidth]{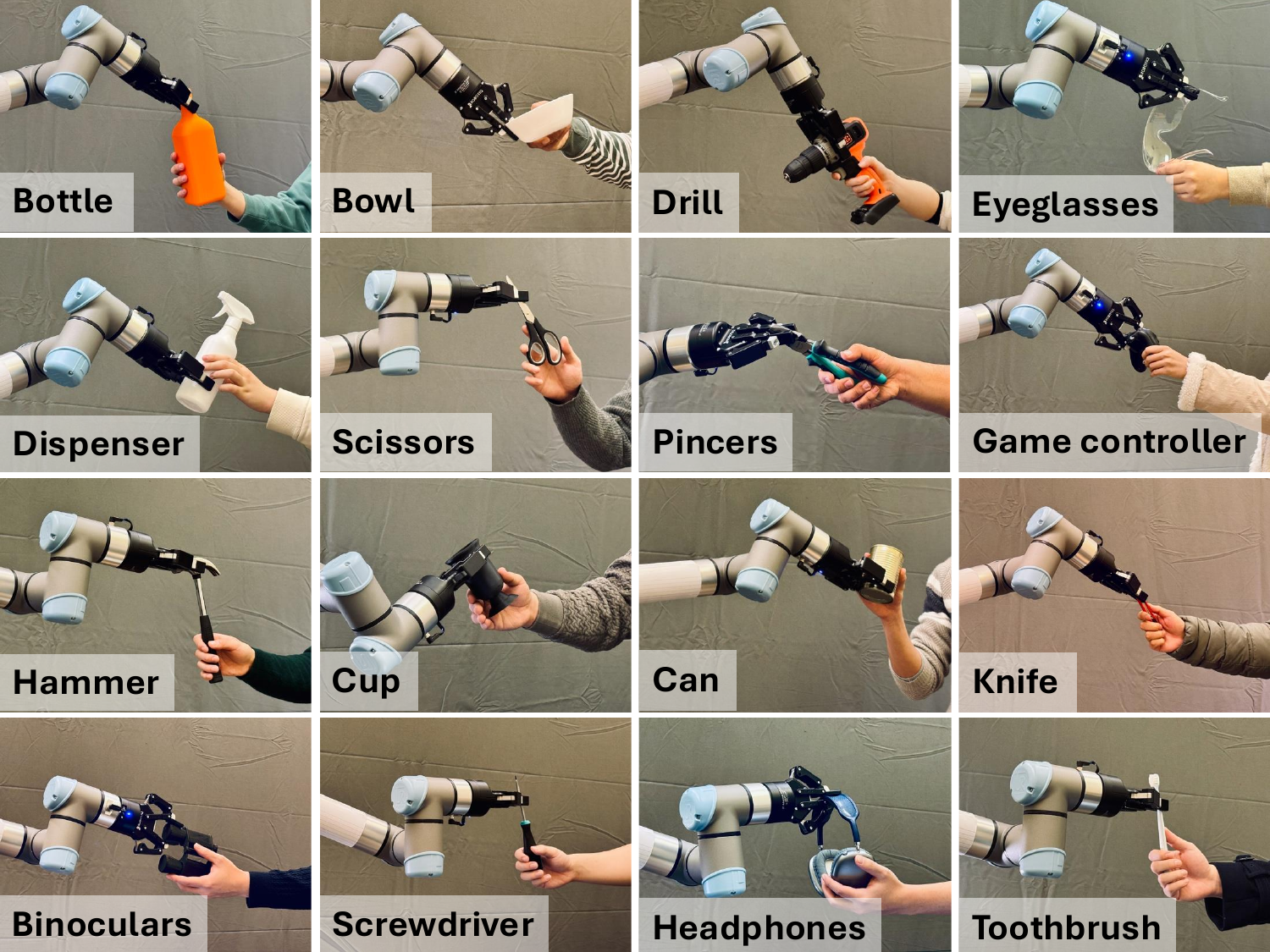}
    \caption{
    \textbf{Representative Full-Strategy Handover Cases.}
    Each example shows an object selected from the user's spoken intent, grasped with human-usage region awareness, and delivered to the user's hand.
    Cases span multiple object categories and intent expressions, providing qualitative support for the objective results in TABLE~\ref{tab:fs_performance}.
    }
    \label{fig:handover_cases}
\end{figure}

\subsection{Effect of Hand-Gripper Collision Avoidance on Safety and Comfort (RQ2)}
 \label{subsec:results_rq2}
To answer RQ2, we compared the full strategy (\textbf{FS}) against \textbf{A2} and \textbf{A3} using Bonferroni-adjusted pairwise tests (TABLE~\ref{tab:pairwise_comparison}). 
In the \textbf{FS} vs \textbf{A2} contrast, disabling hand-gripper collision avoidance sharply reduces \emph{perceived safety} ($p<.001$) while \emph{interaction comfort} does not change ($p=.16$), meaning people feel less safe but the handover can still feel similarly comfortable. In the \textbf{FS} vs \textbf{A3} contrast, \textbf{FS} is rated higher on both \emph{safety} ($p<.001$) and \emph{comfort} ($p<.01$), meaning comfort drops most when both hand-gripper collision avoidance and human-usage region awareness are absent. 
These findings partially support \textbf{H2}: disabling hand-gripper collision avoidance alone significantly reduces perceived safety, but comfort does not decrease until human-usage region awareness is also removed (A3).
This is likely because people can adjust their receiving pose to compensate, preserving comfort; however, the perceived risk of collision still reduces safety. Comfort drops only when the grasp additionally makes the object awkward to receive.

\begin{table}[t]
\centering
\caption{Objective performance of \textbf{Intent-Handover (FS)} over 150 trials (30 participants, 5 trials each): pipeline success rates and mean processing times per stage.}
\label{tab:fs_performance}
\setlength{\tabcolsep}{2.5pt}
\renewcommand{\arraystretch}{1.16}
\footnotesize
\begin{tabular*}{\columnwidth}{@{\extracolsep{\fill}}p{0.30\columnwidth}p{0.49\columnwidth}c@{}}
\toprule
\textbf{Category} & \textbf{Metric} & \textbf{Value} \\
\midrule
\multicolumn{3}{@{}l@{}}{\colorbox{gray!15}{\parbox{\dimexpr\columnwidth-2\fboxsep\relax}{\textbf{Success Rates}}}} \\
Intent identification & Correct object and grasp region inferred from natural language & 88.00\% \\
Handover execution & Robot delivered object to user's hand (given correct intent) & 83.33\% \\
Overall pipeline & Identification correct \& handover successful & 73.33\% \\
\midrule
\multicolumn{3}{@{}l@{}}{\colorbox{gray!15}{\parbox{\dimexpr\columnwidth-2\fboxsep\relax}{\textbf{Average Time}}}} \\
Reasoning stage & Mean intent reasoning time & 2.47\,s \\
Execution stage & Mean handover execution time & 8.83\,s \\
Overall pipeline & Mean total time & 11.30\,s \\
\bottomrule
\end{tabular*}
\end{table}
\subsection{Full-Strategy Pipeline Performance}
TABLE~\ref{tab:fs_performance} summarizes direct logs for \textbf{FS} over 150 trials (30 participants, 5 FS trials each), without questionnaire scoring.
Intent identification succeeded in 88.00\% of trials (132/150), handover execution succeeded in 83.33\% of correctly identified trials (110/132), and the combined overall pipeline achieved a success rate of 73.33\% (110/150).
Intent identification failures mainly occurred when multiple objects in the scene were semantically compatible with the user's instruction (e.g., several objects could plausibly match the described use), causing the VLM to select an unintended target.
Handover execution failures were primarily caused by the selected grasp pose leading to infeasible motion plans for the robot arm.
Mean reasoning time was 2.47\,s, mean execution time was 8.83\,s, and mean total pipeline time was 11.30\,s.
Fig.~\ref{fig:handover_cases} provides representative \textbf{FS} handover cases as qualitative support for these objective logs.



\section{CONCLUSIONS}
We presented \textbf{Intent-Handover}, a robot-to-human handover system that performs handovers directly from natural language via vision--language intent inference and grasp optimization. 
Our user study indicates that human-usage region awareness mainly improves trust through perceived competence and reliability, hand-gripper collision avoidance mainly improves perceived safety, and comfort is best preserved when both are enabled. 
Future work will extend the system to richer language interaction, broader objects and users, and additional robot platforms.

\bibliographystyle{ieeetr}
\bibliography{reference}

\end{document}